\documentclass[10pt,journal]{IEEEtran}
\ifCLASSOPTIONcompsoc
\else
\fi

\ifCLASSINFOpdf
  \usepackage[pdftex]{graphicx}
  \graphicspath{{../pdf/}{../jpeg/}}
  \DeclareGraphicsExtensions{.pdf,.jpeg,.png}
\else
  \usepackage[dvips]{graphicx}
  \graphicspath{{../eps/}}
  \DeclareGraphicsExtensions{.eps}
\fi

\usepackage{algorithm}
\usepackage{algpseudocode}
\usepackage{amsmath}
\usepackage{amssymb}
\usepackage{graphics}
\usepackage{epsfig}
\usepackage{times}
\usepackage{graphicx}
\usepackage{color}
\usepackage{url}
\usepackage{multirow}
\usepackage{threeparttable}

\begin{document}
%
\title{Deep Visual Attention Prediction}
\author{Wenguan Wang, and Jianbing Shen,~\IEEEmembership{Senior Member,~IEEE}
\IEEEcompsocitemizethanks{%
\IEEEcompsocthanksitem This work was supported in part by the National Basic Research Program of China (973 Program) (No. 2013CB328805),
the National Natural Science Foundation of China (No. 61272359), and the Fok Ying-Tong Education Foundation for Young Teachers.
Specialized Fund for Joint Building Program of Beijing Municipal Education Commission. (Corresponding author: \textit{Jianbing Shen})
\IEEEcompsocthanksitem W. Wang and J. Shen are with Beijing Laboratory of Intelligent Information Technology,
School of Computer Science, Beijing Institute of Technology, Beijing 100081, P. R. China.
(Email: \{wenguanwang, shenjianbing\}@bit.edu.cn)
}
\thanks{}
}
\markboth{IEEE Transactions on Image Processing} 
{Shell \MakeLowercase{\textit{et al.}}: Bare Demo of IEEEtran.cls
for Computer Society Journals}

\maketitle
\begin{abstract}
In this work, we aim to predict human eye fixation with
view-free scenes based on an end-to-end deep learning architecture.
Although Convolutional Neural Networks (CNNs) have made substantial improvement on human attention prediction, it is still needed
to improve CNN based attention models by efficiently leveraging multi-scale features.
Our visual attention network is proposed to capture hierarchical saliency information from deep,
coarse layers with global saliency information to shallow, fine layers with local saliency response.
Our model is based on a skip-layer network structure, which predicts human attention from multiple convolutional layers with various reception fields.
Final saliency prediction is achieved via the cooperation of those global and local predictions.
Our model is learned in a deep supervision manner, where supervision is directly fed into
multi-level layers, instead of previous approaches of providing supervision only at the output layer and
propagating this supervision back to earlier layers.
Our model thus incorporates multi-level saliency predictions within a single network, which significantly decreases the redundancy of previous approaches of learning multiple network streams with different input scales. Extensive experimental analysis on various challenging benchmark datasets demonstrate
our method yields state-of-the-art performance with competitive inference time\footnote{Our source
code is available at \url{https://github.com/wenguanwang/deepattention}.}.
\end{abstract}

\begin{IEEEkeywords}
Visual attention, convolutional neural network, saliency detection, deep learning, human eye fixation.
\end{IEEEkeywords}

%

\IEEEpeerreviewmaketitle

\section{Introduction}
\label{section1}
Humans have astonishing ability to quickly pay attention to parts of the image instead of processing the whole scene in its entirety. Simulating such selective attention mechanism of Human Visual System (HVS), which commonly referred as \textit{visual attention prediction} or \textit{visual saliency detection}\footnote{In this paper, the terms attention, saliency, and eye fixation are used interchangeably.}, is a classic research area
in the fields of computer vision and neuroscience. This modeling not only gives an insight into HVS, but also shows much potential in areas such as image cropping, object recognition \cite{gao2004discriminant}, visual tracking \cite{mahadevan2009saliency}, object segmentation \cite{wang2015saliency,shen2014lrw,wang2015videosalient}, video understanding \cite{zhang2017revealing,Yang2017a,Yang2017b,Yang2017c,zhang2015shadow}, to name a few.

In the past few decades, many models have been developed to quantitatively predict human eye attended locations in the form of a \textit{saliency map}, where a brighter pixel indicates higher probability of gaining human attention. These models are generally classified into two main categories as bottom-up approaches \cite{le2006coherent,gao2008discriminant} and  top-down approaches \cite{gao2009discriminant,kanan2009sun,borji2012probabilistic}. The former methods are stimulus-driven, which infer the human attention based on visual stimuli themselves without the knowledge of the image content. In contrast, the top-down  attention mechanisms are task-driven and usually require explicit understanding of the context of scene. Learning with a specific class is therefore a
frequently adopted principle.

\begin{figure}
  \centering
      \includegraphics[width=0.99 \linewidth]{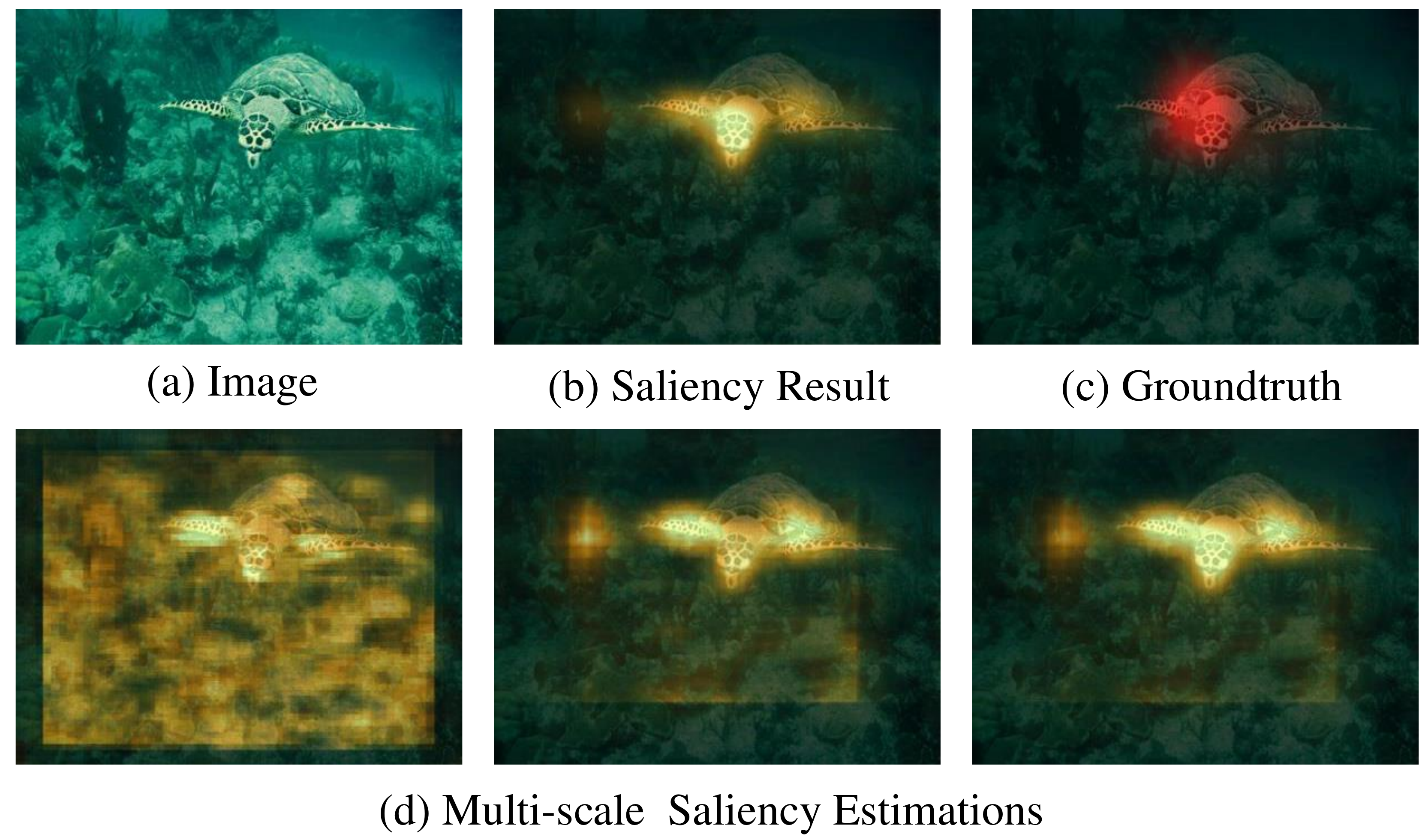}
\caption{The proposed attention model efficiently infers human attention (b) via incorporating multi-scale and multi-level saliency information (d) from different convolution layers within a single network. }
\label{fig0}
\end{figure}

Early bottom-up attention models \cite{le2006coherent,gao2008discriminant} mainly adopted hand-designing features (\textit{e.g.}, intensity,
color, and edge orientation) or heuristics (\textit{e.g.}, center-surround contrast \cite{gao2008discriminant}) based on limited human knowledge on visual
attention. Recently, it has observed a new wave of development \cite{vig2014large,liu2016learning,jetley2016end,pan2016shallow,kruthiventi2016saliency,fang2016learning,wang2017videosalient} using \textit{Convolutional Neural Networks} (CNNs) that emphasizes the importance of automatic hierarchical feature extraction and end-to-end task learning. Provided with
enough training data, deep learning architectures have been shown impressive performance on a
diverse set of visual tasks, ranging from a global scale image classification
\cite{krizhevsky2012imagenet}, to a more local object detection \cite{girshick2014rich} or semantic segmentation \cite{long2015fully}.

In this work, we address the problem of task-free bottom-up visual attention of predicting human eye fixations in natural images with a CNN architecture.
CNNs are powerful visual models of learning features from data and yield hierarchies of features by building high-level features from low-level ones.
It is also well-known that hierarchical processing is ubiquitous in low-level HVS \cite{hubel1962receptive}.
This makes CNN models a powerful tool for tackling the problem of human eye fixation prediction. For fully exploiting the powerful hierarchical representations of CNNs, a skip architecture is
designed to capture multi-level saliency response, ranging from the local to the global, using
shallow to deep convolutional layers with small to large receptive fields. The proposed CNNs based attention model learns visual attention at
multiple scales and multiple levels in a deep supervision manner \cite{lee2015deeply}.
As shown in Fig. \ref{fig0}, the final attention prediction is achieved via the deep fusion of various saliency estimations from multiple levels.
Another advantage is that the multi-scale saliency is learned within a single network, which is more succinct compared with conventional multi-scale attention models with multi-stream networks.

The core trainable network of our attention model works with an \textit{encoder-decoder} architecture, where the encoder network is topologically identical to the first 13 convolutional layers in the VGG16 network \cite{simonyan2014very} and decoder network is for mapping the low resolution encoder feature maps dense full-input-resolution feature maps. The decoder performs upsampling with inverse convolutions, which is also termed as \textit{deconvolution}, and also achieves dimensionality reduction for compressing the encoder feature maps. The upsampling is performed with
trainable multi-channel upsampling kernels, which is more favored than previous attention methods with a fixed bilinear interpolation kernel.  The attention model is trained using whole images and corresponding ground truth
saliency maps. When testing, saliency predictions can be generated by directly feed-forwarding testing images,
without relying on any prior knowledge. The proposed saliency model inherits the advantages of \textit{Fully Convolutional Network} \cite{long2015fully} that utilizes multi-layer information for pixel-wise prediction.

We comprehensively evaluate our method on the five public challenging datasets: MIT300 \cite{judd2012benchmark}, MIT1003 \cite{judd2009learning}, TORONTO \cite{bruce2006saliency}, PASCAL-S \cite{li2014secrets} and DUT-OMRON \cite{yang2013saliency}, where the proposed attention model produces more accurate saliency maps than state-of-the-arts. Meanwhile, it achieves a frame rate of 10fps on a GPU. Thus it is a practical attention prediction model in terms of both speed and accuracy. To summarize, the main contributions are three-fold:
\begin{itemize}
\item We investigate convolutional neural networks for saliency prediction, which captures multi-level saliency information within a single network.
      It is designed to be efficient both in terms of memory and computational time.
\item The proposed model is trained with deep supervision manner, which feeds supervision directly into multiple layers, thus naturally learns multi-level saliency information and improves the discriminativeness and robustness of learned saliency features.
\item The proposed model works in an encoder-decoder architecture, where the decoder performs upsampling with
trainable multi-channel kernels. The effectiveness is confirmed by comparisons with others in experiments.
\end{itemize}

\section{Related work}
\label{section2}
In this section, we first give a brief review of related research for saliency detection. Then we summarize the typical deep learning architectures used in saliency detection.

\subsection{Saliency Detection}
\label{section2.1}
Traditional saliency algorithms with a long history targeted at \textit{visual attention prediction}, which refers to
the task of identifying the fixation points that human viewers would focus
on at first glance. The work of Itti \textit{et al.} \cite{itti1998model}, which was inspired by the Koch and Ullman model \cite{koch1987shifts},
was one of the earliest computational models. Since then, many follow-up works \cite{gao2004discriminant} have been proposed in this direction.
In recent decades, there is a new wave in saliency detection  \cite{li2015visual,wang2016st,gong2015saliency,Fu2015Normalized,liu2014,liu2014s,liu2016,Ye2017,wang2017saliency,wang2017videosalient}  that concentrated on uniformly highlighting the most salient object regions in an image, starting with the works of \cite{liu2007learning} and \cite{achanta2009frequency}. The later methods, also named as \textit{salient object detection}, are directly driven by
object-level tasks. In this study, we mainly overview the typical works of the first type of saliency models, since our method tries to predict human eye fixations over an image. We refer the reader to recent literatures (\cite{borji2013state} and \cite{borji2015salient}) for more detailed overviews.

\begin{figure*}[t]
  \centering
      \includegraphics[width=0.99 \linewidth]{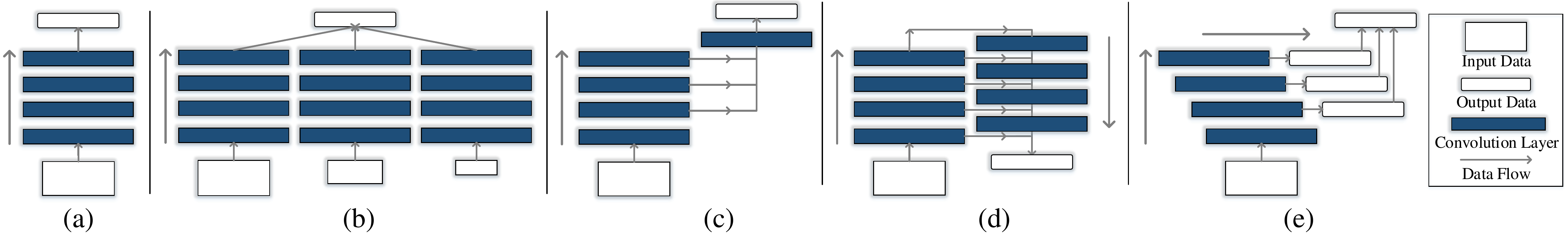}
\caption{(a)-(c) Illustration of three typical categories of deep learning architectures used in attention prediction: (a) single-stream network architecture, (b) multi-stream network architecture with multi-scale inputs, and (c) skip-layer network architecture. (d) Bottom-up/top-down network architecture used in salient object segmentation and instance segmentation. (e) The adopted network is a skip-layer network, which incorporates multi-scale saliency information within a single network. It is capable of learning robust features via integrating supervision into both earlier and last layers. See Section~\ref{section2.2} for more details.}
\label{fig1}
\end{figure*}

Most of classic attention models belong to bottom-up mechanism. The basis of those bottom-up models can date back to Treisman and Gelade's~\cite{treisman1980feature} Feature Integration Theory (FIT), where they combined important visual features to direct human attention over pop-out and conjunction search tasks. Typical attention models consist of three cascaded components: visual feature extraction, saliency inference, and saliency integration. Low-level features, \textit{e.g.}, intensity, color, and orientation \cite{itti1998model}, are first engineered by hand. Inspired
by studies that the salient regions in the visual field would first pop out through different low-level features from their surroundings, \textit{center-surround contrast} is widely adopted for
inferring the saliency. Then saliency is either computed by the relative difference between a region and its local surrounding \cite{itti1998model,harel2006graph,judd2009learning},  or calculating global rarity of features over the entire scene \cite{bruce2006saliency,zhang2008sun}.  Since saliency is computed over several features in parallel, the final step is for fusing them in a scalar map called the ``saliency map''. This step is guided by different principles, \textit{e.g.}, pre-defined linear weights \cite{itti1998model}, trainable weights based on Support Vector Machine (SVM) \cite{judd2009learning}. From the view of mechanism to obtain attention, previous models can also be classified as \cite{borji2013state}: cognitive model \cite{itti1998model,le2006coherent}, Bayesian model \cite{zhang2008sun}, decision theoretic model \cite{gao2004discriminant,gao2009discriminant}, information theoretic model \cite{bruce2006saliency}, graphical model \cite{harel2006graph,liu2007learning},  spectral analysis model \cite{achanta2009frequency}, pattern classification model \cite{judd2009learning} and some other models \cite{goferman2012context} that are based on other mechanism.

In the last few years, many deep learning architectures have been proposed for object recognition. Those deep learning solutions generally achieved better performance, compared with traditional non-deep learning techniques. The Ensemble of Deep Networks (eDN) \cite{vig2014large} represented an early architecture that automatically learns the representations for saliency prediction, blending feature maps from different layers. DeepGaze \cite{Kummerer2014b} fed the responses of different layers of AlexNet \cite{krizhevsky2012imagenet} and a predefined center bias into a softmax layer, and generated a probability distribution of human eye fixation. DeepGaze II \cite{kummerer2016deepgaze} (from DeepGaze \cite{Kummerer2014b}), recently was proposed that provides a deeper network with VGG19 \cite{simonyan2014very}, where the attention information is directly inferred from the original VGGNet, without fine-tuning on attention dataset.  Kruthiventi \textit{et al.} \cite{kruthiventi2015deepfix} also proposed a DeepFix model based on VGG16. In \cite{kruthiventi2016saliency}, saliency prediction and object detection were achieved in a deep CNN. SALICON net \cite{huang2015salicon} captured multi-scale saliency via concatenating fine and coarse features from two stream convolutional networks trained with multi-scale inputs.
In \cite{pan2016shallow}, two models with shallow and deep network architectures were exploited for saliency prediction. Jetley \textit{et al.} \cite{jetley2016end} tested several loss functions based on probability distance measures, such as $\chi^2$ divergence, total variation distance, cosine distance, KL divergence and Bhattacharyya distance by considering saliency models as generalized Bernoulli distributions.

\subsection{Deep Learning Architectures of Saliency Detection Models}
\label{section2.2}
In this section, we discuss typical deep learning architectures of previous deep saliency models and present a graphical illustration in Fig.~\ref{fig1}.  We classify the configurations
of exiting deep saliency models into three main categories: i) \textit{single-stream network}; ii) \textit{multi-stream network} learning with multi-scale inputs; and iii) \textit{skip-layer network}. For completeness, we also include an architecture, namely \textit{bottom-up/top-down network}, which is used in salient object segmentation and instance segmentation.
Clarifying these alternative architectures would help make clearer the advantages of our adopted network with respect to previous methods.

\subsubsection{Single-stream network}
As demonstrated in Fig. \ref{fig1}(a), single-stream network is the standard architecture of CNNs based attention models, which is opted by many saliency detection works \cite{kruthiventi2015deepfix,jetley2016end,kruthiventi2016saliency,pan2016shallow}. All other kinds of deep learning architectures can be viewed as variations of single-stream network.
It has been proved that saliency cues on different level and scales are important in saliency detection~\cite{yang2013saliency}. Incorporating multi-scale feature representations of neural
networks  into attention models is a natural choice. In the following variation of single-stream network, namely multi-stream network, the modifications are performed in this line.

\subsubsection{Multi-stream network}
Examples of this form of network include \cite{huang2015salicon,zhao2015saliency,li2015visual,liu2016learning}. The key concept in multi-stream network is illustrated in Fig. \ref{fig1}(b). This kind of network pursues learning multi-scale saliency information via training multiple  networks with  multi-scale inputs. The multiple network streams are parallel and may have different architectures, corresponding to multiple scales.
As demonstrated in \cite{xie2015holistically}, input data are simultaneously fed into multiple
streams, and then the feature responses from various streams are concatenated and fed into a global output
layer to produce the final saliency.
We can find that, with multi-stream network, the multi-scale learning happens ``outside'' the networks. In the following architecture, the multi-scale or multi-level learning is ``inside'' the network, which is achieved via combining hierarchical features from multiple convolutional layers.

\subsubsection{Skip-layer network}
A typical skip-layer learning architecture is shown in Fig. \ref{fig1}(c), which is adopted in \cite{Kummerer2014b,kummerer2016deepgaze,cornia2016deep}.
Rather than training multiple parallel streams on multiple scaled input images, skip-layer network learns multi-scale features ``inside''  a primary stream.
Multi-scale responses are learned from different layers with increasingly larger receptive fields and downsampling ratios,
and these responses are then concatenated together for outputting final saliency.

\subsubsection{Bottom-up/top-down network}
Readers may also be interested in a recent network architecture, called bottom-up/top-down network, which is used in salient object segmentation \cite{Liu_2016_CVPR} and instance segmentation \cite{pinheiro2016learning}. We show the configuration of such network in Fig. \ref{fig1}(d), where segmentation features are first generated via traditional bottom-up convolution manner, and then a top-down refinement is proceeded for merging the information from deep to shallow layers into segmentation mask. The main rationale behind this architecture is to generate high-fidelity object masks since deep convolutional layers should lose detailed image information. The bottom-up/top-down network can be seen as a kind of skip-layer network, as different layers are also linked together.
\begin{figure*}[t]
  \centering
      \includegraphics[width=0.99 \linewidth]{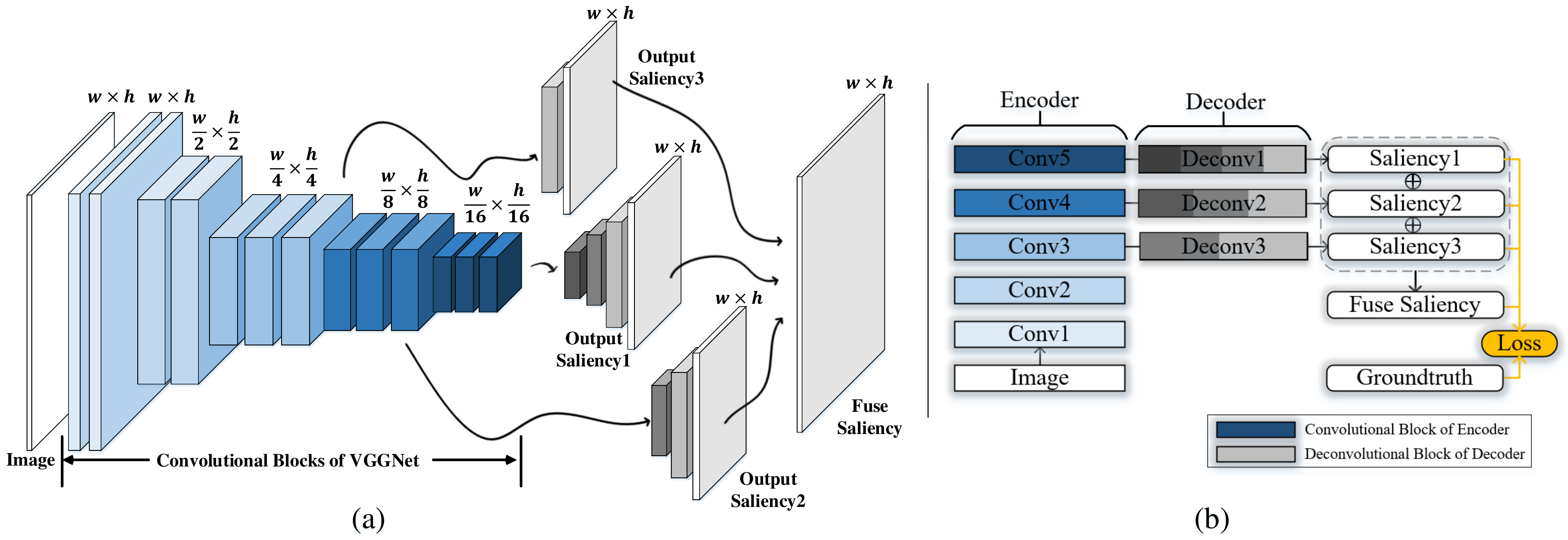}
\caption{Architecture of our deep attention model. (a) Our attention model learns to combine multi-level saliency information from different layers with various receptive field sizes. (b) The proposed deep attention network adopts an encoder-decoder architecture. Moreover, the supervision is directly fed into hidden layers, encouraging the model to learn robust features and generate multi-scale saliency estimates. See Section \ref{section3} for more details.}
\label{fig2}
\end{figure*}

\subsubsection{The adopted network}
We show the architecture of our attention model in Fig. \ref{fig1}(e), which is inspired by the network in \cite{xie2015holistically} and deeply-supervised network in \cite{lee2015deeply}. The network incorporates multi-scale and multi-level attention information from different layers, which is learned in a deeply supervised manner. The major difference between the adopted network and previous models is that, our network provides integrated direct supervision to the hidden layers, rather than the standard approach of providing supervision only at the output layer and propagating this supervision back to earlier layers. The multi-level and multi-scale saliency is explicitly learned from different layers with corresponding supervision. Such hidden layer supervision brings improvement in both performance and robustness of features, as discussed in \cite{lee2015deeply,xie2015holistically}.

It inherits the advantage of skip-layer network (Fig. \ref{fig1}(c)) that does not require learning multiple networks streams with multi-scale inputs.
It is also a light-weighted version compared with multi-stream network (Fig. \ref{fig1}(d)) and bottom-up/top-down network (Fig. \ref{fig1}(e)). We find the bottom-up/top-down network is difficult to train in practice while the network equipped with deep supervision gains high training efficiency.

\section{Our approach}
\label{section3}
\subsection{Architecture Overview}
CNNs are capable of capturing the hierarchy of features, where the
lower layers respond to primitive image features such as edges, corners
and shared common patterns, and the higher layers extract semantic
information like object parts or faces. Such low and high-level features are shown
to be both important and complementary in estimating visual
attention, which motivates us incorporates multi-layer information together for inferring the visual attention.

The architecture of the proposed model is depicted in Fig. \ref{fig2}. Multi-scale predictions are learned from different layers with different receptive field sizes (see Fig. \ref{fig2}(a)). For obtaining such multi-scale predictions, supervision is directly fed into corresponding layers (see Fig. \ref{fig2}(b)). Such deep supervision learning strategy boosts the performance via: 1) directly producing multi-scale saliency predictions; and 2) improving discriminativeness of intermediate layers, thus gaining improvement of overall performance, as demonstrated in  \cite{lee2015deeply}.

For recovering the spatial information destroyed by the pooling operation in the convolutional layers, our model works in an encoder-decoder architecture. The encoder part captures high-level features via convolving and downsampling the low-level feature map, which decreases the size of the feature maps from bottom to up. Our decoder network upsamples feature maps, which constructs an output that maintains the original resolution of the input. The decoder part also brings two advantages: 1) the convolutional filters used in decoder network are learnable, which is preferable to the fixed interpolation kernel used in previous methods; and 2) the decoder gradually reduces feature dimensions for higher computation efficiency.

\subsection{Proposed Attention Model}
The proposed attention model is a fully convolutional neural network, which is trained
to predict pixel-wise saliency values for a given image in an end-to-end manner.
Our encoder part is a stack of convolutional layers.
A convolutional layer is defined on a translation invariance basis and shared weights across different spatial locations.
Each layer input and output in a convolutional network are three-dimensional tensors, called feature maps.
The first layer is the image, with pixel size $h$ and $w$, and three channels.
The output feature map is obtained by convolving the input feature map with a linear filter, then adding a bias term.
For improving translation invariance and representation capability, convolutional layers are usually interleaved with non-linear down-sampling operation (\textit{e.g.}, max pooling) and point-wise nonlinearity (\textit{e.g.}, \textit{ReLU}s).
If we denote the input feature map of $l$-th layer as $\textbf{X}^{l-1}$, whose convolution filters are determined by the weights $\textbf{W}^l$, then the output feature map $\textbf{X}^{l}$ of $l$-th layer is obtained via:
\begin{equation}
    \begin{aligned}
     &\textbf{X}^{l} = f_{con}(\textbf{X}^{l-1};\textbf{W}^{l}_{con}) = \textbf{W}^{l}_{con} \ast \textbf{X}^{l-1}, l = 1...L\\
     &\textbf{X}^{0} = \textbf{I},
    \end{aligned}
    \label{eq:1}
\end{equation}
where $\ast$ denotes the convolution operation, $L$ denotes the total number of layers, $\textbf{X}^{0}$ is the input image $\textbf{I}$. For simplicity, in our present discussion, we absorb the bias term into the weight parameters and omit the activate and max-pooling operation.

Due to the stride of convolutional and feature pooling layers, detailed spatial information is lost, thus the local output feature maps are very coarse. For upsampling the coarse feature map, \textit{deconvolution} (transposed convolution) layer could be adopted.
Deconvolution layer works by swapping the forward and backward passes of a convolution, which upsamples the input feature map via backwards convolution:
\begin{equation}
    \begin{aligned}
     &f_{decon}(\textbf{X};\textbf{W}_{decon}) = \textbf{W}_{decon} \circledast_s \textbf{X},\\
    \end{aligned}
    \label{eq:2}
\end{equation}
where the $\circledast_s$ indicates fractionally strided convolution, which can be viewed as the reverse convolution operation via adding stride $s$ into the input. The stride $s$ is a upsampling factor. Activation functions can also be attached for learning a nonlinear upsampling \cite{long2015fully}.
For restoring downsampled feature map $\textbf{X}^{l}$ of $l$-th layer to a fine feature map with the same size as input, a decoder with multiple deconvolution layers could be added on the top of $\textbf{X}^{l}$:
\begin{equation}
    \begin{aligned}
    &\textbf{Y}^{l} = D(\textbf{X}^{l};\textbf{W}^{l}_{decon}),
    \end{aligned}
    \label{eq:3}
\end{equation}
where the $D$ indicates a set of deconvolution operations and the $W^{l}_{decon}$ indicates all the kernel weights of the deconvolution layers.
Again, the non-linear activation layers are omitted. Then a classifier, composed of a $1\times1$ convolution layer with \textit{sigmoid} nonlinearity, is added to produce saliency map with the same size of the input image.

In the encoder network, several convolutional layers and pooling layers are
stacked alternately in depth, thus hierarchical features are extracted
with increasingly larger receptive fields. In this way, the low level features are characterized via lower layers, while high-level semantic features are encoded in higher layers.

Previous works have shown that saliency is best captured
when features are considered from multiple scales. This motivates us select $M$ layers from the encoder network for explicitly predicting saliency in multi-scales and multi-levels, where each selected layer is
associated with a decoder network. Thus we are able to obtain $M$ output attention prediction maps with the same size of the input. For combining all the parameters of the convolution and deconvolution layers of the encoder and decoder networks, we define:
\begin{equation}
    \begin{aligned}
    \textbf{W} = \{\textbf{W}^1_{con}, \ldots, \textbf{W}^L_{con}, \textbf{W}^{l_1}_{decon}, \ldots, \textbf{W}^{l_M}_{decon}\}.
    \end{aligned}
    \label{eq:4}
\end{equation}
For each decoder network, the
weights of the corresponding classifier are denoted as $\textbf{w}_c^m$. Then we also combine all the parameters of classifiers together:
\begin{equation}
    \begin{aligned}
    \textbf{w}_c = \{\textbf{w}_c^1, \ldots, \textbf{w}_c^M\}.
    \end{aligned}
    \label{eq:5}
\end{equation}
We derive an objective function that merges all the output-layer classification error:
\begin{equation}
    \begin{aligned}
    \mathcal{Q}(\textbf{W},\textbf{w}_c) = \sum_{m=1}^M\mathcal{L}(\textbf{W},\textbf{w}_c^m),
    \end{aligned}
    \label{eq:6}
\end{equation}
where the $\mathcal{L}$ denotes the image-level loss function for saliency prediction. Given an image $\textbf{I}$ with size $h \times w \times 3$ and its groundtruth attention map $\textbf{G}\in [0,1]^{h \times w}$, $\mathcal{L}$ is defined as the cross-entropy loss:
\begin{equation}
    \begin{aligned}
    \mathcal{L}(\textbf{W},\textbf{w}_c^m)=\!-\!\!\sum\nolimits_{i=1}^{|\textbf{I}|}&\big(\textbf{G}_i\log P(\textbf{S}^m_i=1|\textbf{I},\textbf{W},\textbf{w}_c^m)\\
    &+\!(1\!-\!\textbf{G}_i)\log P(\textbf{S}^m_i\!=\!0|\textbf{I},\textbf{W},\textbf{w}_c^m)\big),
    \end{aligned}
    \label{eq:7}
\end{equation}
where $\textbf{S}^m$ indicates the predicted attention map from $m$-th decoder network, and $|\textbf{I}|$ refers to the number of pixels in $\textbf{I}$.

For fusing the multi-layer output saliency predictions, an ``attention fusion'' layer with $1\times1$ convolution kernel is added to merge all the predicted attention maps $\{\textbf{S}^m\}_{m=1}^{M}$, which simultaneously
learns the fusion weight during training. Let $F$ indicate the fused attention prediction: $\textbf{F} = \sum_{m=1}^M w^m_{f}\textbf{S}^m$ and $\textbf{w}_{f} = \{w^m_{f}\}_{m=1}^{M}$ is the fusion weight,
the loss function for the fusion layer is defined as:
\begin{equation}\small
    \begin{aligned}
    \!\!\!\!\mathcal{P}(\textbf{W}, \textbf{w}_c, \textbf{w}_{f})\!=\!\!-\!\!\sum\nolimits_{i=\!1}^{|\textbf{I}|}&\big(\textbf{G}_i\log P(\textbf{F}_i=1|\textbf{I},\textbf{W},\textbf{w}_c,\textbf{w}_{f}) \\&+\!(1\!-\!\textbf{G}_i)\!\log P(\textbf{F}_i\!=\!0|\textbf{I},\textbf{W},\textbf{w}_c,\textbf{w}_{f})\big),
    \end{aligned}
    \label{eq:8}
\end{equation}
Then all the parameters $\textbf{W}$, $\textbf{w}_{c}$, and $\textbf{w}_{f}$ can be learned via minimizing the following objective
function over all the training set via standard (back-propagation) stochastic gradient
descent:
\begin{equation}
    \begin{aligned}
    \!\!(\textbf{W}, \textbf{w}_{c}, \textbf{w}_{f})^*\!=\!argmin(\frac{1}{M}\mathcal{Q}(\textbf{W},\textbf{w}_{c})\!+\!\mathcal{P}(\textbf{W}, \textbf{w}_{c}, \textbf{w}_{f})).
    \end{aligned}
    \label{eq:9}
\end{equation}
After training, given a test image, we can use the trained CNN model to predict a pixel-level attention map.

\subsection{Implementation Detail}
\subsubsection{Encoder Network}
The encoder part of our network is inspired by the VGG16 \cite{simonyan2014very} that consists of five convolutional blocks and three fully connected layers.
Since our network explicitly utilizes the extracted CNNs feature maps, we only consider convolutional
layers and omit the fully connected layers. In the standard VGG16 model, with an input image having a size of $h \times w \times 3$,  the spatial dimensions of the features generated from the last convolution layer (\textit{conv5-3}) is $\frac{h}{32} \times \frac{w}{32}$ which is relatively small for the saliency prediction task. For preserving more spatial information of the feature map, we modify the network
structure via removing the final pooling layer (\textit{pool5}). This results in an output feature blob of spatial dimensions $\frac{h}{16} \times \frac{w}{16}$ after the last convolution layer.

The units of the CNNs are sensitive to small sub-regions of the visual field, called a receptive field. The receptive field of deeper convolution layer with respect to the input image is larger with the convolution and pooling operations. Therefore, in the encoder network, stacking many convolution layers leads to gradually learning ``local'' to ''global'' saliency information (\textit{i.e.}, responsive to increasingly larger region of pixel space). For capturing multi-scale saliency information, we select $M = 3$ feature maps generated respectively from \textit{conv3-3}, \textit{conv4-3}, and \textit{conv5-3} convolution layers of our encoder network. Then those saliency maps are fused for inferring the final saliency prediction. We empirically find that considering further more layers does not contribute to performance improvement, but brings extra computation burden.

\subsubsection{Decoder Network}
For each saliency feature map, a decoder with multiple deconvolution layers is added to gradually enlarge the spatial dimension until obtaining saliency prediction with original input size. The saliency feature map from \textit{conv4-3} layer of the encoder network, for example, has spatial size of $\frac{h}{8} \times \frac{w}{8}$. Since spatial dimensions of the output blob are halved after each convolution block,
its decoder network has three deconvolution layers, where each deconvolution layer doubles the spatial size of input feature correspondingly. Thus the spatial dimensions of the features gradually increase as $\{\frac{h}{8} \times \frac{w}{8}\}\!\rightarrow\!\{\frac{h}{4} \times \frac{w}{4}\}\!\rightarrow\!\{\frac{h}{2} \times \frac{w}{2}\}\!\rightarrow\!\{h \times w\}\!$. Each deconvolution layer is equipped with \textit{ReLU} layer, which learns a nonlinear upsampling. Analogously, the decoder networks of \textit{conv3-3} layer and \textit{conv5-3} layer have two and four deconvolution layers, respectively.

Starting from the first convolutional block, the number of channels in the outputs of successive blocks gradually increase as $64\rightarrow\!128\!\rightarrow\!256\!\rightarrow\!512\!\rightarrow\!512$. This enables the net to progressively learn richer semantic representations of the input image. However, maintaining the channel dimension of the feature map unchanged within the decoder network will cause large redundancy both in terms of memory and computational time during inference. Therefore, our decoder network not only increases the spatial dimension of the feature map, but also reduces the dimensionality of the feature channel space. Again, taking the output of \textit{conv4-3} layer as an example, the channel dimension is decreased as $256\!\rightarrow\!128\!\rightarrow\!64\!\rightarrow\!32\!\rightarrow\!1$ via three deconvolution layers and final classifier with $1\times1$ convolution layer and \textit{sigmoid} activity function.

\subsubsection{Training and Testing}
The proposed deep saliency network
is implemented with the \textit{Caffe} library \cite{jia2014caffe}.
While training, the weights of the filters in the five
convolution blocks of the encoder network are initialized
from the VGG16, which is trained on the \textit{ImageNet} \cite{krizhevsky2012imagenet} database
for the task of classification. The weights of the remaining layers are
randomly initialized from a Gaussian distribution with zero mean and standard deviation of 0.01.
We train the networks on the 10,000 images from the
\textit{SALICON} \cite{jiang2015salicon} training set where eye fixation annotations
are simulated through mouse movements of users on blurred images.
The authors of \cite{jiang2015salicon} show that the mouse-contingent
saliency annotations strongly correlate with actual eye-tracker annotations.
A mini-batch of 16 images is used in each iteration.
The learning rate is initialized as $1\times10^{-4}$ and scaled down by a factor of 0.1 after 2,000 iterations.
The network was validated against the \textit{SALICON} validation set (5,000 images) after every 100 iterations to monitor
convergence and overfitting.
The network parameters are learned by back-propagating the loss function in (\ref{eq:9})
using Stochastic Gradient Descent (SGD) with momentum.
During testing, given a query image, we obtain
final saliency prediction from the last multi-scale attention fusion layer.  Our model achieves
processing speed as little as 0.1 seconds on our PC with a TITANX GPU and 32G RAM.

\section{Experimental Results}
\label{section4}
\subsection{Datasets}
We conducted evaluation on five widely used saliency datasets with different characteristics.

\subsubsection{MIT300 \cite{judd2012benchmark}}The \textit{MIT300} dataset\footnote{Available at:\url{http://saliency.mit.edu/results_mit300.html}} is a collection of 300
natural images where saliency maps were generated from
eye-tracking data of 39 users. Saliency maps of this entire
dataset are held out.

\subsubsection{MIT1003 \cite{judd2009learning}}The \textit{MIT1003} dataset\footnote{Available at:\url{http://people.csail.mit.edu/tjudd/WherePeopleLook}} contains 1,003 images from Flickr and LabelMe, including
779 landscape and 228 portrait images. The groundtruth saliency maps have
been created from eye-tracking data of 15 human observers.

\subsubsection{TORONTO \cite{bruce2006saliency}}The \textit{TORONTO} dataset\footnote{Available at:\url{http://www-sop.inria.fr/members/Neil.Bruce}} is one of the most widely used dataset. It contains 120 color images with
resolution of $511 \times 681$ pixels from indoor and outdoor environments.
Images are presented at random to 20 subjects
for 3 seconds with 2 seconds of gray mask in between.
\subsubsection{PASCAL-S \cite{li2014secrets}}The \textit{PASCAL-S} dataset\footnote{Available at:\url{http://cbi.gatech.edu/salobj/}} uses the 850 natural images of the validation set of the PASCAL
VOC 2010 \cite{everingham2010pascal}, with the eye fixations
during 2 seconds of 8 subjects.

\subsubsection{DUT-OMRON \cite{yang2013saliency}} The \textit{DUT-OMRON} dataset\footnote{Available at:\url{http://saliencydetection.net/dut-omron/}} consists of 5,168 images
with the largest height or width of 400 pixels. Each image is viewed by 5 subjects;
then, a postprocessing step is applied to remove outlier eye
fixation points that do not lie on a meaningful object.

\begin{table}
\caption{CHARACTERISTICS OF 5 SELECTED EYE-TRACKING DATASETS.
}\label{table1}
\centering
\renewcommand\arraystretch{1.1}
\begin{tabular}{|c||c|c|c|}  
\hline
 &\#Images &\#Viewers &Resolution\\
\hline
\hline
MIT300 \cite{judd2012benchmark}        &300 	&39 	&$\max(w,h) = 1024$\\
MIT1003 \cite{judd2009learning}	       &1,003 	&15 	&$\max(w,h) = 1024$\\
TORONTO \cite{bruce2006saliency}       &120 	&20 	&$511\times681$\\
PASCAL-S \cite{li2014secrets}           &850 	&8 	&$\max(w,h) = 500$\\
DUT-OMRON \cite{yang2013saliency}      &5,168 	&5 	&$\max(w,h) = 400$\\
\hline
\end{tabular}
\end{table}

Some statistics and features of these datasets
are summarized in Table \ref{table1}. Above datasets record human eye fixation positions with eye-tracking equipment. Once the human eye fixations are collected, they often convert discrete fixations into a continuous distribution,
a ground-truth saliency map, by smoothing. Each fixation location is blurred with a Gaussian kernel, whose parameters are established by the authors.

\subsection{Evaluation Metrics}
There are several ways to measure the agreement between
model predictions and human eye fixations. Previous studies on saliency metrics \cite{riche2013saliency} show that it's hard to achieve a fair comparison for
evaluating saliency models by any single metric. Here, we carried out our quantitative experiments by comprehensively considering a variety of different metrics, including Earth Mover¡¯s Distance (EMD),
Normalized Scanpath Saliency (NSS), Similarity Metric (SIM), Linear Correlation
Coefficient (CC), AUC-Judd, AUC-Borji, and shuffled AUC. Those metrics are selected since they are widely-accepted and standard for evaluating a saliency model.

\begin{table}
\caption{Evaluation Metrics.
}\label{table2}
\renewcommand\arraystretch{1.1}
\setlength\tabcolsep{2pt}
\centering
\begin{tabular}{|c||c|c|}  
\hline
 &Category &Groundtruth \\
\hline
\hline
Earth Mover¡¯s Distance (EMD)       &Distribution-based 	&Saliency Map $G$	\\
Linear Correlation Coefficient (CC)       &Distribution-based 	&Saliency Map $G$	\\
Similarity Metric (SIM)           &Distribution-based 	&Saliency Map $G$	\\
Normalized Scanpath Saliency (NSS)	       &Value-based 	&Fixation Map $Q$	\\
AUC-Judd      &Location-based 	&Fixation Map 	Q\\
AUC-Borji     &Location-based 	&Fixation Map Q	\\
shuffled AUC      &Location-based 	&Fixation Map Q	\\
\hline
\end{tabular}
\end{table}

For simplification, we denote the predicted saliency map
as $S$, the map of fixation locations as $Q$ and the continuous saliency map
(distribution) as $G$. In Table \ref{table2}, we list the characteristics formation of our adopted evaluation metrics.
Next we describe these evaluation metrics in detail.

\subsubsection{Earth Mover¡¯s Distance (EMD)}
EMD measures the distance between two 2D maps, $\textbf{G}$ and $\textbf{S}$.
It is the minimal cost of transforming the probability distribution
of the estimated saliency map $\textbf{S}$ to that of the ground truth
map $\textbf{G}$. Therefore, a low EMD corresponds to a high-quality saliency map.

\subsubsection{Normalized Scanpath Saliency (NSS)}
NSS is a metric specifically designed for saliency map evaluation.
Given saliency map $\textbf{S}$ and a binary map of fixation locations $\textbf{Q}$:
\begin{equation}
    \begin{aligned}
    &NSS = \frac{1}{N}\sum\nolimits_{i=1}^N\overline{\textbf{S}}(i)\times \textbf{Q}(i),\\
    &where ~~N = \sum\nolimits_i\textbf{Q}(i)~~ and ~~\overline{\textbf{S}} = \frac{\textbf{S}-\mu(\textbf{S})}{\sigma(\textbf{S})},
    \end{aligned}
\end{equation}
where $N$ is the total number of human eye positions and $\sigma(\cdot)$ stands for standard deviation. This metric is calculated by taking the mean of scores assigned by the unit normalized saliency map (with zero mean and unit standard deviation) at human eye fixations.

\begin{table}
\caption{CHARACTERISTICS OF THE PROPOSED MODEL AND 13
SATE-OF-THE-ART SALIENCY MODELS.
}\label{table3}
\renewcommand\arraystretch{1.1}
\setlength\tabcolsep{2pt}
\centering
\begin{tabular}{|c||c|c|c|c|}  
\hline
Model &Input Size &Training &Deep Learning &Runtime\\
\hline
\hline
DeeFix \cite{kruthiventi2015deepfix}       &- 	&Yes 	&Yes &-\\
SALICON \cite{huang2015salicon}	       &$\max\{w,h\}$=800 	&Yes 	&Yes &-\\
Mr-CNN \cite{liu2016learning}       &$400\times400$ 	&Yes 	&Yes & 14s*\\
SalNet \cite{pan2016shallow}           &$320\times240$ 	&Yes 	&Yes & 0.1s*\\
Deep Gaze I	\cite{Kummerer2014b}      &full size	&Yes 	&Yes &-\\
SU \cite{kruthiventi2016saliency}     &$417\times417$ &Yes 	&Yes &-\\
eDN	 \cite{vig2014large}     &$512\times384$ 	&Yes 	&Yes &8s*\\
AIM \cite{bruce2009saliency}            &$\frac{1}{2}$full size 	&Yes 	&No &2s\\
Judd Model	\cite{judd2009learning}     &$200\times200$ 	&Yes 	&No &10s\\
BMS	 \cite{zhang2013saliency}     &$w=600$ 	&No 	&No &0.3s\\
CAS	\cite{goferman2012context}      &$\max\{w,h\}$=250 	&No 	&No &16s\\
GBVS	 \cite{harel2006graph}     &full size 	&No 	&No &2s\\
ITTI \cite{itti1998model}     &full size 	&No 	&No &4s\\
\hline
\hline
DVA &$\max\{w,h\}$=256 	&Yes 	&Yes &0.1s*\\
\hline
\end{tabular}
\leftline{}
\leftline{~~~~-~The authors have not released detailed information.}
\leftline{~~~~*~Runtime with GPU.}
\end{table}

\subsubsection{Linear Correlation Coefficient (CC)}
The CC is a statistical method generally measures how correlated or dependent two variables are.
CC can be used to interpret saliency and fixation maps, $\textbf{G}$ and $\textbf{S}$ as
random variables to measure the linear relationship
between them:
\begin{equation}
    \begin{aligned}
    CC = \frac{cov(\textbf{S},\textbf{G})}{\sigma(\textbf{S})\times\sigma(\textbf{G})},
    \end{aligned}
\end{equation}
where $cov(\textbf{S},\textbf{G})$ is the covariance of $\textbf{S}$ and $\textbf{G}$. It ranges
between -1 and +1, and a score close to -1 or +1 indicates
a perfect alignment between the two maps.

\subsubsection{Similarity Metric (SIM)}
The SIM measures the similarity between
two distributions, viewed as histograms. SIM is computed as
the sum of the minimum values at each pixel, after
normalizing the input maps:
\begin{equation}
    \begin{aligned}
    &SIM = \sum\nolimits_{i=1}\min(\textbf{S}'(i),\textbf{G}'(i)),\\
    &where ~~\sum\nolimits_i\textbf{S}'(i)=1~~ and ~~\sum\nolimits_i\textbf{G}'(i)=1,
    \end{aligned}
\end{equation}
where $\textbf{S}'$ and $\textbf{G}'$ are normalized to be probability distributions, given a saliency map $\textbf{S}$ and the continuous fixation map $\textbf{G}$.
A SIM of one indicates the distributions are the same, while a SIM of zero indicates no overlap.

\subsubsection{AUC \cite{bruce2006saliency}}
The AUC metric, defined as the area under the receiver operating characteristic (ROC) curve,
is widely used to evaluate the maps by saliency models.
Given an image and its groundtruth eye fixation
points, fixated points and other ones are regarded as the
positive and negative sets, respectively. Then, the computed saliency map is binarily
classified into salient region and non-salient region by using a
threshold. Through varying the threshold from 0 to 1, ROC curve is
obtained by plotting true positive rate versus false positive rate,
with its underneath area calculated as AUC score.
AUC can be greatly influenced by center-bias and border cut.

Depending upon the choice of the non-fixation
distribution, there are several variants of AUC.
In our experiments, we opt AUC-Judd, AUC-Borji and the shuffled AUC (s-AUC).
The former two variants choose non-fixation points with a uniform
distribution, while the last one, shuffled AUC, uses human fixations of
other images as non-fixation distribution.

\begin{table}
\caption{COMPARISON OF QUANTITATIVE SCORES OF DIFFERENT SALIENCY MODELS on MIT300 \cite{judd2012benchmark} Dataset.
}\label{table4}
\centering
\resizebox{0.48\textwidth}{!}{
\renewcommand\arraystretch{1.1}
\setlength\tabcolsep{1pt}
\begin{tabular}{|c||c|c|c|c|c|c|c|}  
\hline
Model &AUC-Judd $\uparrow$ &SIM $\uparrow$ &EMD $\downarrow$ &AUC-Borji $\uparrow$ &s-AUC $\uparrow$ &CC $\uparrow$ &NSS $\uparrow$\\
\hline
\hline
Humans                                              &0.92 	&1.00 	&0.00 	&0.88 	&0.81 	&1.00 	&3.29 	\\
\hline
\hline
DeeFix\cite{kruthiventi2015deepfix}	            &\textbf{0.87} 	&\textbf{0.67} 	&\textbf{2.04} 	&0.80 	&0.71 	&\textbf{0.78} 	&\textbf{2.26} 	\\
SALICON\cite{huang2015salicon}                     &\textbf{0.87} 	&0.60 	&2.62 	&\textbf{0.85} 	&\textbf{0.74} 	&0.74 	&2.12 \\
Mr-CNN\cite{liu2016learning}                      &0.77 	&0.45 	&4.33 	&0.76 	&0.69 	&0.41 	&1.13\\
SalNet\cite{pan2016shallow}                        &0.83 	&0.51 	&3.35 	&0.82 	&0.65 	&0.55 	&1.41\\
Deep Gaze I\cite{Kummerer2014b}                    &0.84 	&0.39 	&4.97 	&0.83 	&0.66 	&0.48 	&1.22\\
BMS\cite{zhang2013saliency}                       &0.83 	&0.51 	&3.35 	&0.82 	&0.65 	&0.55 	&1.41 \\
eDN\cite{vig2014large}                            &0.82 	&0.41 	&4.56 	&0.81 	&0.62 	&0.45 	&1.14 \\
CAS\cite{goferman2012context}                     &0.74 	&0.43 	&4.46 	&0.73 	&0.65 	&0.36 	&0.95 \\
AIM \cite{bruce2009saliency}                        &0.77 	&0.40 	&4.73 	&0.75 	&0.66 	&0.31 	&0.79 \\
Judd Model\cite{judd2009learning}                 &0.81 	&0.42 	&4.45 	&0.80 	&0.60 	&0.47 	&1.18\\
GBVS\cite{harel2006graph}                      &0.81 	&0.48 	&3.51 	&0.80 	&0.63 	&0.48 	&1.24\\
ITTI\cite{itti1998model}                       &0.75 	&0.44 	&4.26 	& 0.74	&0.63 	&0.37 	&0.97\\
\hline						
\hline
DVA                                            &0.85   &0.58   &3.05   &0.78   &0.71    &0.68   &1.98\\
\hline
\end{tabular}
}
\end{table}

\begin{table}
\caption{COMPARISON OF QUANTITATIVE SCORES OF DIFFERENT SALIENCY MODELS on  MIT1003 \cite{judd2009learning} Dataset.
}\label{table5}
\centering
\resizebox{0.48\textwidth}{!}{
\renewcommand\arraystretch{1.1}
\setlength\tabcolsep{2pt}
\begin{tabular}{|c||c|c|c|c|c|c|}  
\hline
Model &AUC-Judd $\uparrow$ &SIM $\uparrow$  &AUC-Borji $\uparrow$ &s-AUC $\uparrow$ &CC $\uparrow$ &NSS $\uparrow$\\
\hline
\hline
Mr-CNN	\cite{liu2016learning}                      &0.80 	        &0.35 	 	     &0.77  &0.73 	&0.38 	&1.36\\
SU \cite{kruthiventi2016saliency}                   &- 	            &- 	     	     &-      &0.71 	&- 	     &2.08\\
DeeFix* \cite{kruthiventi2015deepfix}	            &0.90* 	        &0.54* 	 &0.87*      &0.74* 	&0.72* 	&2.58*\\
BMS	 \cite{zhang2013saliency}                       &0.79 	        &0.33 	 	     &0.76 &0.69 	&0.36 	&1.25\\
eDN	 \cite{vig2014large}                            &0.85 	        &0.30 	 	     &0.84 &0.66 	&0.41 	&1.29\\
CAS	\cite{goferman2012context}                      &0.76  	        &0.32  	  	     &0.74 &0.68  	&0.31  	&1.07 \\
AIM \cite{bruce2009saliency}                        &0.79  	        &0.27  	  	     &0.76 &0.68 	&0.26 	& 0.82\\
Judd Model	\cite{judd2009learning}                 & 0.76          &0.29  	  	     &0.74 &0.68 	&0.30  	&1.02 \\
GBVS	 \cite{harel2006graph}                      &0.83 	        &0.36 	 	     &0.81 &0.66 	&0.42 	&1.38\\
ITTI	 \cite{itti1998model}                       &0.77 	        &0.32 	 	     &0.76 &0.66 	&0.33 	&1.10\\
\hline
\hline
DVA                                                 &\textbf{0.87}           &\textbf{0.50}            &\textbf{0.85}  &\textbf{0.77}    &\textbf{0.64}   &\textbf{2.38}\\
\hline
\end{tabular}
}
\leftline{}
\leftline{*~DeeFix \cite{kruthiventi2015deepfix} is fine-tuned with 900 images of MIT1003 dataset, while}
\leftline{~~~evaluated on the other 103 images.}
\end{table}

\begin{table}
\caption{COMPARISON OF QUANTITATIVE SCORES OF DIFFERENT SALIENCY MODELS on TORONTO \cite{bruce2006saliency} Dataset.
}\label{table6}
\centering
\resizebox{0.49\textwidth}{!}{
\renewcommand\arraystretch{1.1}
\setlength\tabcolsep{2pt}
\begin{tabular}{|c||c|c|c|c|c|c|}  
\hline
Model &AUC-Judd $\uparrow$ &SIM $\uparrow$  &AUC-Borji $\uparrow$ &s-AUC $\uparrow$ &CC $\uparrow$ &NSS $\uparrow$\\
\hline
\hline
Mr-CNN	\cite{liu2016learning}                      &0.80 	&0.47 	&0.79 	&0.71 	&0.49 	&1.41\\
eDN	 \cite{vig2014large}                            &0.85 	&0.40 	&0.84 	&0.62 	&0.50 	&1.25\\
CAS	\cite{goferman2012context}                      &0.78 	&0.44 	&0.78 	&0.69 	&0.45 	&1.27\\
AIM \cite{bruce2009saliency}                        &0.76   &0.36   &0.75    &0.67    &0.30   &0.84\\
Judd Model	\cite{judd2009learning}                 &0.78 	&0.40 	&0.77	&0.67 	&0.41 	&1.15\\
GBVS	 \cite{harel2006graph}                      &0.83 	&0.49 	&0.83 	&0.64 	&0.57 	&1.52\\
ITTI	 \cite{itti1998model}                       &0.80 	&0.45	&0.80 	&0.65 	&0.48 	&1.30\\
\hline
\hline
DVA                              &\textbf{0.86} 	&\textbf{0.58} 	&\textbf{0.86}    &\textbf{0.76} 	&\textbf{0.72} 	&\textbf{2.12}\\
\hline
\end{tabular}
}
\end{table}

\subsection{Comparison Results}
To demonstrate the effectiveness of the proposed
deep attention model in predicting eye fixations, we evaluated it
by comparison to 13 state-of-the-art models, including six classical models:
ITTI \cite{itti1998model}, GBVS \cite{harel2006graph}, Judd Model \cite{judd2009learning},
BMS	\cite{zhang2013saliency}, CAS \cite{goferman2012context}, AIM \cite{bruce2009saliency},
and seven deep learning based models:
DeeFix \cite{kruthiventi2015deepfix}, SALICON \cite{huang2015salicon}, Mr-CNN \cite{liu2016learning}, SalNet \cite{pan2016shallow},
Deep Gaze I \cite{Kummerer2014b},  eDN \cite{vig2014large}, and SU \cite{kruthiventi2016saliency}.
These methods have been proposed in recent years or widely used for comparison.
For the methods: ITTI \cite{itti1998model}, CAS \cite{goferman2012context}, and AIM \cite{bruce2009saliency}
that we calculated saliency maps using their publicly available code with the recommended parameters by the authors.
A summary of these models is provided in Table \ref{table3}. As seen, most of the models
require off-line training or based on deep learning framework. Our model DVA (\textit{Deep Visual Attention}),
is also included in Table \ref{table3}. We report the computation time of above models with TORONTO \cite{bruce2006saliency} dataset.
Since most deep learning based models have not been publicly available,
we only report the speed performance of Mr-CNN \cite{liu2016learning}, SalNet \cite{pan2016shallow},
and eDN \cite{vig2014large} with other non-deep learning model.
As visible, the suggested model, DVA, achieves a fast speed (0.1s per frame).

The quantitative results obtained on the MIT300 \cite{judd2012benchmark}, MIT1003 \cite{judd2009learning}, TORONTO \cite{bruce2006saliency}, PASCAL-S \cite{li2014secrets} and DUT-OMRON \cite{yang2013saliency} datasets
are presented in Table \ref{table4}, Table \ref{table5}, Table \ref{table6}, Table \ref{table7} and Table \ref{table8}, respectively. In MIT300 dataset (see Table \ref{table4}), our model is struggle to compete with DeeFix \cite{kruthiventi2015deepfix} and SALICON \cite{huang2015salicon} models. DeepFix introduces more complex network architecture and considers center prior. Besides, it is trained with more samples (SALICON dataset + 2700 extra images with actual eye fixation data). SALICON model is fine-tuned with more complex objective functions and images from OSIE dataset \cite{xu2014predicting}. Our method achieves promising results across a wide of datasets, which verifies its robustness and generality. With a relatively lightweight architecture, our work is meaningful for inspiring future work in this direction and offers a deep insight into the advantage of multiscale saliency cues for visual attention prediction.

\begin{table}
\caption{COMPARISON OF QUANTITATIVE SCORES OF DIFFERENT SALIENCY MODELS on PASCAL-S \cite{li2014secrets} Dataset.
}\label{table7}
\centering
\resizebox{0.49\textwidth}{!}{
\renewcommand\arraystretch{1.1}
\setlength\tabcolsep{2pt}
\begin{tabular}{|c||c|c|c|c|c|c|}  
\hline
Model &AUC-Judd $\uparrow$ &SIM $\uparrow$ &AUC-Borji $\uparrow$ &s-AUC $\uparrow$ &CC $\uparrow$ &NSS $\uparrow$\\
\hline
\hline
SU \cite{kruthiventi2016saliency}                   &- 	&- 	&- 	&0.73 	 	&- 	&2.22\\
BMS	 \cite{zhang2013saliency}                       &- 	&- 	&- 	&\textbf{1.32} 	 	&- 	&1.28\\
eDN	 \cite{vig2014large}                            &- 	&- 	&- 	&1.29 	 	&- 	&1.42\\
CAS	\cite{goferman2012context}                      &0.78   &0.34       &0.75 	&0.67    &0.36   &1.12\\
AIM \cite{bruce2009saliency}                        &0.77   &0.30       &0.75 	&0.65    &0.32   &0.97\\
GBVS	 \cite{harel2006graph}                      &0.84 	&0.36 	 	&0.82 	&0.65 	&0.45 	&1.36\\
ITTI	 \cite{itti1998model}                       &0.82 	&0.36 	 	&0.80 	&0.64 	&0.42 	&1.30\\
\hline
\hline
DVA                                                 &\textbf{0.89}   &\textbf{0.52}    &\textbf{0.85} 	&0.77     &\textbf{0.66}   &\textbf{2.26}\\
\hline
\end{tabular}
}
\end{table}

\begin{table}
\caption{COMPARISON OF QUANTITATIVE SCORES OF DIFFERENT SALIENCY MODELS on DUT-OMRON \cite{yang2013saliency} Dataset.
}\label{table8}
\resizebox{0.49\textwidth}{!}{
\centering
\renewcommand\arraystretch{1.1}
\setlength\tabcolsep{2pt}
\begin{tabular}{|c||c|c|c|c|c|c|c|}  
\hline
Model &AUC-Judd $\uparrow$ &SIM $\uparrow$  &AUC-Borji $\uparrow$ &s-AUC $\uparrow$ &CC $\uparrow$ &NSS $\uparrow$\\
\hline
\hline
SU \cite{kruthiventi2016saliency}                   &- 	&- 	&- &0.83 	&- 	&3.02\\
BMS	 \cite{zhang2013saliency}                       &- 	&- 	&- &0.79 	&- 	&1.66\\
eDN	 \cite{vig2014large}                            &- 	&- 	&- &0.80 	&- 	&1.33\\
CAS	\cite{goferman2012context}                      &0.80   &0.37   &0.79 &0.73    &0.40   &1.47\\
AIM \cite{bruce2009saliency}                        &0.77   &0.32   &0.75 &0.69    &0.30   &1.05\\
GBVS	 \cite{harel2006graph}                      &0.87 	&0.43 	&0.85 &0.81 	&0.53 	&1.71\\
ITTI	 \cite{itti1998model}                       &0.83 	&0.39 	&0.83 &0.78 	&0.46 	&1.54\\
\hline
\hline
DVA                                                 &\textbf{0.91}   &\textbf{0.53}   &\textbf{0.86} &\textbf{0.84}    &\textbf{0.67}   &\textbf{3.09}\\
\hline
\end{tabular}
}
\end{table}

\begin{figure*}
  \centering
      \includegraphics[width=0.99 \linewidth]{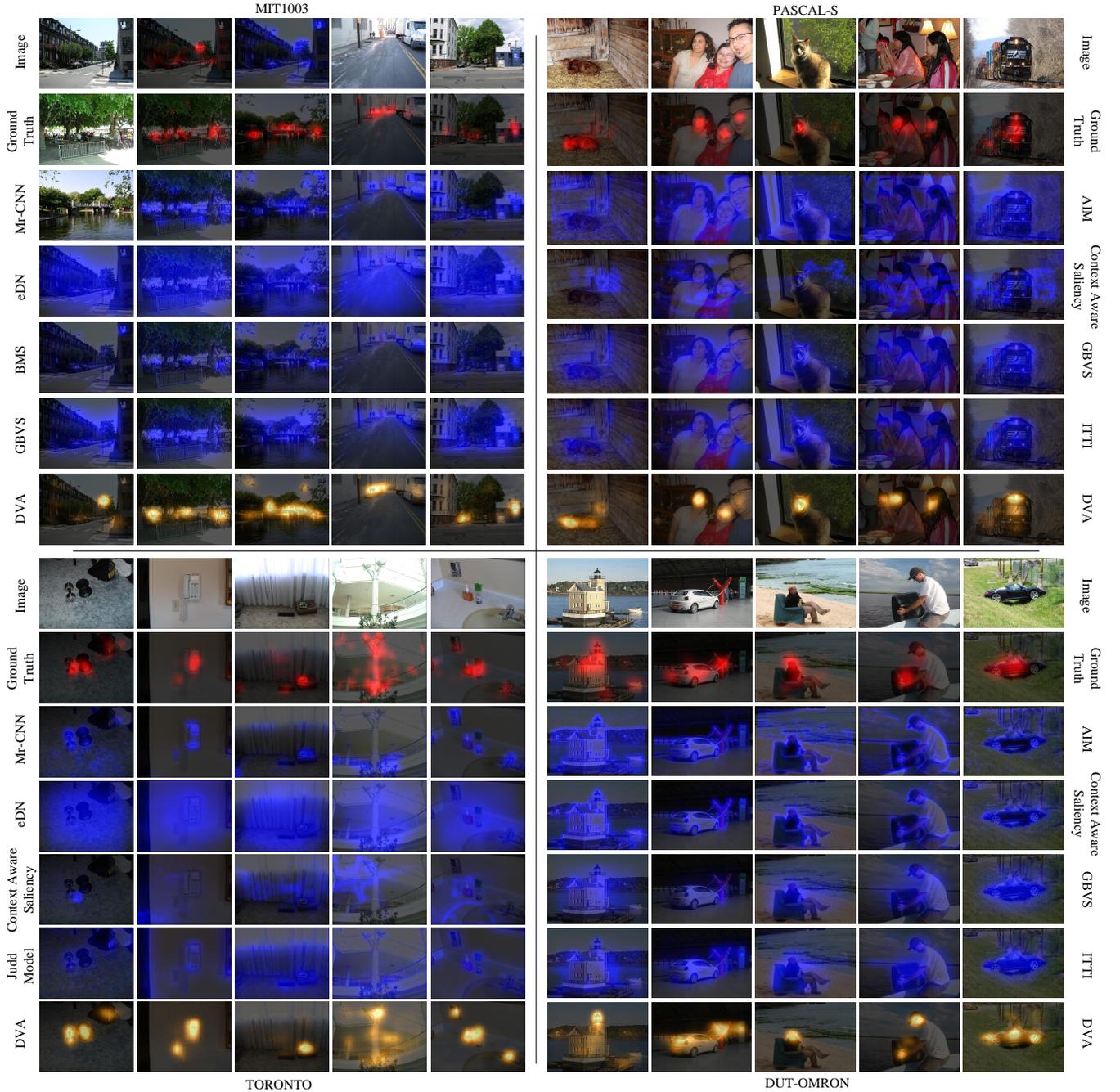}
\caption{Comparison of saliency maps with various state-of-the-art methods on MIT1003, TORONTO \cite{bruce2006saliency},
PASCAL-S \cite{li2014secrets} and  DUT-OMRON \cite{yang2013saliency}  datasets.
Note that the proposed method generates more reasonable saliency maps compared with the state-of-the-art.}
\label{fig3}
\end{figure*}

The qualitative results obtained by the proposed deep attention network,
along with that of other recent methods on a few example
images from MIT1003 \cite{judd2009learning}, TORONTO \cite{bruce2006saliency},
PASCAL-S \cite{li2014secrets} and DUT-OMRON \cite{yang2013saliency}
datasets are presented in Fig. \ref{fig3}. As shown in
the figures, the proposed attention model is able to consistently
capture saliency arising from both low-level features such as colour contrast as well as
the more high-level aspects such
as humans, faces and text. Our saliency maps are very localized
in the salient regions compared with other methods, even when images
have cluttered backgrounds or salient regions in different sizes. We attribute the performance
of the proposed deep attention model to its large depth and the utilization of multi-level features.

\begin{table}
\centering
\caption{Ablation study on TORONTO \cite{bruce2006saliency} dataset using the shuffled AUC and CC metrics. }
\renewcommand\arraystretch{1.1}
\setlength\tabcolsep{2pt}
\label{table9}
\begin{tabular}{|c|c|c|c|c|c|}  
\hline
\multirow{2}*{Aspect}
&\multirow{2}*{Variant}
&  \multicolumn{4}{c|}{TORONTO}\\
\cline{3-6}
&&s-AUC $\uparrow$      &$\Delta$s-AUC  &CC $\uparrow$     &$\Delta$CC\\
\hline
\hline
\multirow{1}*{} &whole model                           &\textbf{0.76}      &-  &\textbf{0.72} &-\\
\hline
\multirow{3}*{submodule}&$\textit{conv3-3} $ output    &0.68      &-0.08   &0.57 &-0.15 \\
&$\textit{conv4-3} $ output                            &0.69      &-0.07   &0.65 &-0.07 \\
&$\textit{conv5-3} $ output                            &0.69      &-0.07   &0.69 &-0.03 \\
\hline
\multirow{1}*{fusion} &$\textit{avg.}$ output          &0.72      &-0.04   &0.68    &-0.04 \\
\hline
\multirow{1}*{supervision}
&w/o deep supervision                                  &0.71      &-0.05   &0.68    &-0.04  \\
\hline
\multirow{1}*{upsampling}
&bilinear interpolation kernel                         &0.74      &-0.02   &0.70    &-0.02  \\
\hline
\end{tabular}
\end{table}

\subsection{Ablation Study}
We evaluate the contribution of each component of the proposed deep attention
model on TORONTO \cite{bruce2006saliency} dataset and measure the performance
using the shuffled AUC and CC metrics. We study different variants of our model in several aspects.
Our experiments are summarized in Table \ref{table9}.

\subsubsection{Submodule} We first examine the performance of individual encoder-decoder layers: \textit{conv3-3}, \textit{conv4-3},
and \textit{conv5-3}, which capture saliency information at different scales.
The predictions using different layers are complementary and the overall
architecture can predict better saliency maps.

\subsubsection{Fusion Strategy} We next study the effect of our fusion strategy.
In our attention model, we adopt a convolution layer for merging the multi-layer output saliency predictions,
which automatically learns the fusion weight during training.
To validate the effectiveness of this strategy, we directly average all of the multi-layer outputs ($\textit{avg.}$\! output).
This experiment yields some interesting results.
First, we can find the averaged prediction achieves higher performance,
compared with single-layer outputs. This validates combining predictions from multiple scales
leads to better performance. Secondly, the learned weighted-fusion achieves best performance.
Thus, our attention fusion strategy contributes to the performance gain.

\subsubsection{Supervision}
Here we discuss the role of deep supervision of our attention model.
For examining the contribution of deep supervision, we train our model with weighted-fusion supervision only (minimizing object function $\mathcal{P}$
in (Equ. \ref{eq:8}) only). We can find that our whole model with both considering weighted-fusion supervision and deep supervision of each output layer
(via (Equ. \ref{eq:9})) is more preferable. This observation verifies that the deep supervision directly improves performance.

\subsubsection{Upsampling}
We experiment with different upsampling strategies. In our decoder network,  the
upsampling is performed with  multi-channel trainable upsampling kernels. Here we test the performance of directly feeding features of \textit{conv3-3}, \textit{conv4-3}, and \textit{conv5-3} into classifier layers and upsampling the multi-layer output saliency via a fixed bilinear interpolation kernel. The result in Table \ref{table9} suggests a drop in performance without using learnable kernel.

\section{Discussions and Conclusions}
\label{section5}
In this work, we have proposed a neural network for predicting human eye
fixations on images. The proposed deep attention model inherits the advantages of deeply supervised
nets and fully utilizes the potential
of neural network for representing hierarchy of features to extract saliency information at multiple scales. It builds on top of encoder-decoder architecture, which is
fully convolutional and results a memory and time-efficient model. Experimental results on five eye-tracking datasets
demonstrate the superior performance of our method with respect to other state-of-the-art methods.

%


\begin{thebibliography}{1}
\bibitem{borji2015salient}
A.~Borji, M.-M. Cheng, H.~Jiang, and J.~Li, ``Salient object detection: A
  benchmark,'' \emph{IEEE Trans. on Image Processing}, vol.~24, no.~12,
  pp. 5706--5722, 2015.

 \bibitem{zhang2017revealing}
D.~Zhang, J. Han, L.~Jiang, S.~Ye, and X.~Chang,
``Revealing event saliency in unconstrained video collection,''
\emph{IEEE Trans. on Image Processing}, vol.~26, no.~4,
  pp. 1746--1758, 2017.

\bibitem{wang2016st}
W.~Wang, J.~Shen, L.~Shao, and F.~Porikli, ``Correspondence driven saliency
  transfer,'' \emph{IEEE Trans. on Image Processing}, vol.~25, no.~11,
  pp. 5025--5034, 2016.

\bibitem{liu2016learning}
N.~Liu, J.~Han, T.~Liu, and X.~Li, ``Learning to predict eye fixations via
  multiresolution convolutional neural networks,'' \emph{IEEE Trans. on
  Neural Networks and Learning Systems}, 2016.

\bibitem{gao2004discriminant}
D.~Gao and N.~Vasconcelos, ``Discriminant saliency for visual recognition from
  cluttered scenes,'' in \emph{Advances in Neural Information Processing
  Systems}, 2004, pp. 481--488.

\bibitem{mahadevan2009saliency}
V.~Mahadevan and N.~Vasconcelos, ``Saliency-based discriminant tracking,'' in
  \emph{Proceedings of the IEEE Conference on Computer Vision and Pattern
  Recognition}, 2009, pp. 1007--1013.

\bibitem{wang2015saliency}
W.~Wang, J.~Shen, and F.~Porikli, ``Saliency-aware geodesic video object
  segmentation,'' in \emph{Proceedings of the IEEE Conference on Computer
  Vision and Pattern Recognition}, 2015, pp. 3395--3402.

\bibitem{le2006coherent}
O.~Le~Meur, P.~Le~Callet, D.~Barba, and D.~Thoreau, ``A coherent computational
  approach to model bottom-up visual attention,'' \emph{IEEE Trans. on
  Pattern Analysis and Machine Intelligence}, vol.~28, no.~5, pp. 802--817,
  2006.

\bibitem{gao2008discriminant}
D.~Gao, V.~Mahadevan, and N.~Vasconcelos, ``The discriminant center-surround
  hypothesis for bottom-up saliency,'' in \emph{Advances in Neural Information
  Processing Systems}, 2008, pp. 497--504.

\bibitem{gao2009discriminant}
D.~Gao, S.~Han, and N.~Vasconcelos, ``Discriminant saliency, the detection of
  suspicious coincidences, and applications to visual recognition,'' \emph{IEEE
  Trans. on Pattern Analysis and Machine Intelligence}, vol.~31, no.~6,
  pp. 989--1005, 2009.

\bibitem{kanan2009sun}
C.~Kanan, M.~H. Tong, L.~Zhang, and G.~W. Cottrell, ``{SUN}: Top-down saliency
  using natural statistics,'' \emph{Visual Cognition}, vol.~17, no. 6-7, pp.
  979--1003, 2009.

\bibitem{borji2012probabilistic}
A.~Borji, D.~N. Sihite, and L.~Itti, ``Probabilistic learning of task-specific
  visual attention,'' in \emph{Proceedings of the IEEE Conference on Computer
  Vision and Pattern Recognition}, 2012, pp. 470--477.

\bibitem{vig2014large}
E.~Vig, M.~Dorr, and D.~Cox, ``Large-scale optimization of hierarchical
  features for saliency prediction in natural images,'' in \emph{Proceedings of
  the IEEE Conference on Computer Vision and Pattern Recognition}, 2014, pp.
  2798--2805.

\bibitem{jetley2016end}
S.~Jetley, N.~Murray, and E.~Vig, ``End-to-end saliency mapping via probability
  distribution prediction,'' in \emph{Proceedings of the IEEE Conference on
  Computer Vision and Pattern Recognition}, 2016, pp. 5753--5761.

\bibitem{pan2016shallow}
J.~Pan, E.~Sayrol, X.~Giro-i Nieto, K.~McGuinness, and N.~E. O'Connor,
  ``Shallow and deep convolutional networks for saliency prediction,'' in
  \emph{Proceedings of the IEEE Conference on Computer Vision and Pattern
  Recognition}, 2016, pp. 598--606.

\bibitem{kruthiventi2016saliency}
S.~S. Kruthiventi, V.~Gudisa, J.~H. Dholakiya, and R.~Venkatesh~Babu,
  ``Saliency unified: A deep architecture for simultaneous eye fixation
  prediction and salient object segmentation,'' in \emph{Proceedings of the
  IEEE Conference on Computer Vision and Pattern Recognition}, 2016, pp.
  5781--5790.

\bibitem{fang2016learning}
S.~Fang, J.~Li, Y.~Tian, T.~Huang, and X.~Chen, ``Learning discriminative
  subspaces on random contrasts for image saliency analysis,'' \emph{IEEE
  Trans. on Neural Networks and Learning Systems}, 2016.

\bibitem{krizhevsky2012imagenet}
A.~Krizhevsky, I.~Sutskever, and G.~E. Hinton, ``Imagenet classification with
  deep convolutional neural networks,'' in \emph{Advances in Neural Information
  Processing Systems}, 2012, pp. 1097--1105.

\bibitem{girshick2014rich}
R.~Girshick, J.~Donahue, T.~Darrell, and J.~Malik, ``Rich feature hierarchies
  for accurate object detection and semantic segmentation,'' in
  \emph{Proceedings of the IEEE Conference on Computer Vision and Pattern
  Recognition}, 2014, pp. 580--587.

\bibitem{long2015fully}
J.~Long, E.~Shelhamer, and T.~Darrell, ``Fully convolutional networks for
  semantic segmentation,'' in \emph{Proceedings of the IEEE Conference on
  Computer Vision and Pattern Recognition}, 2015, pp. 3431--3440.

\bibitem{hubel1962receptive}
D.~H. Hubel and T.~N. Wiesel, ``Receptive fields, binocular interaction and
  functional architecture in the cat's visual cortex,'' \emph{The Journal of
  Physiology}, vol. 160, no.~1, pp. 106--154, 1962.

\bibitem{lee2015deeply}
C.-Y. Lee, S.~Xie, P.~W. Gallagher, Z.~Zhang, and Z.~Tu, ``Deeply-supervised
  nets.'' in \emph{AISTATS}, vol.~2, no.~3, 2015, p.~5.

\bibitem{simonyan2014very}
K.~Simonyan and A.~Zisserman, ``Very deep convolutional networks for
  large-scale image recognition,'' \emph{arXiv preprint arXiv:1409.1556}, 2014.

\bibitem{judd2012benchmark}
T.~Judd, F.~Durand, and A.~Torralba, ``A benchmark of computational models of
  saliency to predict human fixations,'' \emph{MIT Technical Report}, 2012.

\bibitem{judd2009learning}
T.~Judd, K.~Ehinger, F.~Durand, and A.~Torralba, ``Learning to predict where
  humans look,'' in \emph{Proceedings of the IEEE International Conference on
  Computer Vision}, 2009, pp. 2106--2113.

\bibitem{bruce2006saliency}
N.~Bruce and J.~Tsotsos, ``Saliency based on information maximization,''
  \emph{Advances in Neural Information Processing Systems}, vol.~18, p. 155,
  2006.

\bibitem{li2014secrets}
Y.~Li, X.~Hou, C.~Koch, J.~M. Rehg, and A.~L. Yuille, ``The secrets of salient
  object segmentation,'' in \emph{Proceedings of the IEEE Conference on
  Computer Vision and Pattern Recognition}, 2014, pp. 280--287.

\bibitem{yang2013saliency}
C.~Yang, L.~Zhang, H.~Lu, X.~Ruan, and M.-H. Yang, ``Saliency detection via
  graph-based manifold ranking,'' in \emph{Proceedings of the IEEE Conference
  on Computer Vision and Pattern Recognition}, 2013, pp. 3166--3173.

\bibitem{itti1998model}
L.~Itti, C.~Koch, and E.~Niebur, ``A model of saliency-based visual attention
  for rapid scene analysis,'' \emph{IEEE Trans. on Pattern Analysis and
  Machine Intelligence}, vol.~20, no.~11, pp. 1254--1259, 1998.

\bibitem{koch1987shifts}
C.~Koch and S.~Ullman, ``Shifts in selective visual attention: towards the
  underlying neural circuitry,'' in \emph{Matters of Intelligence}, 1987, pp.
  115--141.

\bibitem{shen2014lrw}
J. Shen, Y. Du, W. Wang, X. Li, ``Lazy random walks for superpixel segmentation,''
\emph{IEEE Trans. on Image Processing}, vol. 23, no. 4, pp. 1451-1462, 2014.

\bibitem{li2015visual}
G.~Li and Y.~Yu, ``Visual saliency based on multiscale deep features,'' in
  \emph{Proceedings of the IEEE Conference on Computer Vision and Pattern
  Recognition}, 2015, pp. 5455--5463.

\bibitem{gong2015saliency}
C. Gong and D. Tao and  W. Liu and  S. Maybank and  M. Fang and  K. Fu and  J. Yang,
``Saliency propagation from simple to difficult,''
in \emph{Proceedings of the IEEE Conference on Computer Vision and Pattern Recognition}, 2015, pp. 2531-2539.

\bibitem{Fu2015Normalized}
 K. Fu and C. Gong and  I. Y. H. Gu and  J. Yang,
``Normalized cut-based saliency detection by adaptive multi-level region merging,''
in \emph{IEEE Trans. on Image Processing}, vol. 24, no.12, pp. 5671-5683, 2015.

\bibitem{wang2015videosalient}
W. Wang, J. Shen, and L. Shao,
``Consistent video saliency using local gradient flow optimization and global refinement,''
\emph{IEEE Trans. on Image Processing}, vol.~24, no.~11,
  pp. 4185--4196, 2015.

\bibitem{liu2014}
Z. Liu and W. Zou and and O. Le Meur,
``Saliency tree: A novel saliency detection framework,''
in \emph{IEEE Trans. on Image Processing}, vol. 23, no. 5, pp. 1937-1952, 2014.

\bibitem{liu2014s}
Z. Liu and X. Zhang and S. Luo and O. Le Meur,
``Superpixel-based spatiotemporal saliency detection,''
in \emph{IEEE Trans. on Circuits and Systems for Video Technology}, vol. 24, no. 9, pp. 1522-1540, 2014.

\bibitem{liu2016}
Z. Liu and J. Li and L. Ye and G. Sun and L. Shen,
``Saliency detection for unconstrained videos using superpixel-level graph and spatiotemporal propagation,''
in \emph{IEEE Trans. on Circuits and Systems for Video Technology}, 2016.

\bibitem{Ye2017}
L. Ye and Z. Liu and L. Li and L. Shen and C. Bai and Y. Wang,
``Salient object segmentation via effective integration of saliency and objectness,''
in \emph{IEEE Trans. on Multimedia}, 2017.

\bibitem{liu2007learning}
T.~Liu, J.~Sun, N.-N. Zheng, X.~Tang, and H.-Y. Shum, ``Learning to detect a
  salient object,'' in \emph{Proceedings of the IEEE Conference on Computer
  Vision and Pattern Recognition}, 2007, pp. 1--8.

\bibitem{achanta2009frequency}
R.~Achanta, S.~Hemami, F.~Estrada, and S.~Susstrunk, ``Frequency-tuned salient
  region detection,'' in \emph{Proceedings of the IEEE Conference on Computer
  Vision and Pattern Recognition}, 2009, pp. 1597--1604.

\bibitem{borji2013state}
A.~Borji and L.~Itti, ``State-of-the-art in visual attention modeling,''
  \emph{IEEE Trans. on Pattern Analysis and Machine Intelligence},
  vol.~35, no.~1, pp. 185--207, 2013.

\bibitem{treisman1980feature}
A.~M. Treisman and G.~Gelade, ``A feature-integration theory of attention,''
  \emph{Cognitive Psychology}, vol.~12, no.~1, pp. 97--136, 1980.

\bibitem{harel2006graph}
J.~Harel, C.~Koch, P.~Perona \emph{et~al.}, ``Graph-based visual saliency,''
  \emph{Advances in Neural Information Processing Systems}, vol.~1, no.~2,
  p.~5, 2006.

\bibitem{zhang2015shadow}
L.Zhang, Q. Zhang, C. Xiao,
``Shadow remover: image shadow removal based on illumination recovering optimization,''
\emph{IEEE Trans. Image Processing}, vol. 24, no. 11, pp. 4623-4636, 2015.

\bibitem{zhang2008sun}
L.~Zhang, M.~H. Tong, T.~K. Marks, H.~Shan, and G.~W. Cottrell, ``{SUN}: A
  bayesian framework for saliency using natural statistics,'' \emph{Journal of
  Vision}, vol.~8, no.~7, pp. 32--32, 2008.

\bibitem{goferman2012context}
S.~Goferman, L.~Zelnik-Manor, and A.~Tal, ``Context-aware saliency detection,''
  \emph{IEEE Trans. on Pattern Analysis and Machine Intelligence},
  vol.~34, no.~10, pp. 1915--1926, 2012.

\bibitem{Kummerer2014b}
M.~K{\"u}mmerer, L.~Theis, and M.~Bethge, ``Deep gaze {I}: Boosting saliency
  prediction with feature maps trained on imagenet,'' in \emph{Proceedings of
  the International Conference on Learning Representations Workshop}, 2015.

\bibitem{wang2017videosalient}
W. Wang, J. Shen, and L. Shao,
``Video salient object detection via fully convolutional networks,''
\emph{IEEE Trans. on Image Processing}, vol.~27, no.~1, pp. 38--49, 2018.

\bibitem{kummerer2016deepgaze}
M.~K{\"u}mmerer, T.~S. Wallis, and M.~Bethge, ``Deepgaze {II}: Reading
  fixations from deep features trained on object recognition,'' \emph{arXiv
  preprint arXiv:1610.01563}, 2016.

\bibitem{kruthiventi2015deepfix}
S.~S. Kruthiventi, K.~Ayush, and R.~V. Babu, ``Deepfix: A fully convolutional
  neural network for predicting human eye fixations,'' \emph{arXiv preprint
  arXiv:1510.02927}, 2015.

\bibitem{huang2015salicon}
X.~Huang, C.~Shen, X.~Boix, and Q.~Zhao, ``{SALICON}: Reducing the semantic gap
  in saliency prediction by adapting deep neural networks,'' in
  \emph{Proceedings of the IEEE International Conference on Computer Vision},
  2015, pp. 262--270.

\bibitem{Yang2017a}
Y. Yang, C. Deng, S. Gao, W. Liu, D. Tao, X. Gao,
``Discriminative Multi-Instance Multi-Task Learning for 3D Action Recognition,''
 \emph{IEEE Trans. on Multimeddia}, vol. 19, no. 3, pp. 519-529, 2017.

\bibitem{Yang2017b}
Y. Yang, C. Deng, S. Gao, W. Liu, D. Tao, X. Gao,
``Latent Max-Margin Multitask Learning with Skelets for 3D Action Recognition,''
 \emph{IEEE Trans. on Cybernetics}, vol. 47, no. 2, pp. 439-448, 2017.

\bibitem{Yang2017c}
Y. Yang, R. Liu, C. Deng, X. Gao,
``Multi-Task Human Action Recognition via Exploring Super-Category,''
 \emph{Signal Processing}, vol. 124, pp. 36-44, 2016.

\bibitem{wang2017saliency}
W. Wang, J. Shen, R. Yang, and F. Porikli,
``Saliency-aware video object segmentation,''
\emph{IEEE Trans. on Pattern Analysis and Machine Intelligence}, vol.~40, no.~1, pp. 20-33, 2018.

\bibitem{zhao2015saliency}
R.~Zhao, W.~Ouyang, H.~Li, and X.~Wang, ``Saliency detection by multi-context
  deep learning,'' in \emph{Proceedings of the IEEE Conference on Computer
  Vision and Pattern Recognition}, 2015, pp. 1265--1274.

\bibitem{xie2015holistically}
S.~Xie and Z.~Tu, ``Holistically-nested edge detection,'' in \emph{Proceedings
  of the IEEE International Conference on Computer Vision}, 2015, pp.
  1395--1403.

\bibitem{cornia2016deep}
M.~Cornia, L.~Baraldi, G.~Serra, and R.~Cucchiara, ``A deep multi-level network
  for saliency prediction,'' in \emph{Proceedings of the International
  Conference on Pattern Recognition}, 2016.

\bibitem{Liu_2016_CVPR}
X. Yao, J. Han, D. Zhang and F. Nie,
``Revisiting co-saliency detection: a novel approach based on two-stage multi-view spectral rotation co-clustering,''
\emph{IEEE Trans. on Image Processing}, vol. 26, no. 7, pp. 3196-3209, 2017.

\bibitem{pinheiro2016learning}
P.~O. Pinheiro, T.-Y. Lin, R.~Collobert, and P.~Doll{\'a}r, ``Learning to
  refine object segments,'' in \emph{Proceedings of the European Conference on
  Computer Vision}, 2016, pp. 75--91.

\bibitem{jiang2015salicon}
M.~Jiang, S.~Huang, J.~Duan, and Q.~Zhao, ``{SALICON}: Saliency in context,''
  in \emph{Proceedings of the IEEE Conference on Computer Vision and Pattern
  Recognition}, 2015, pp. 1072--1080.

\bibitem{jia2014caffe}
Y.~Jia, E.~Shelhamer, J.~Donahue, S.~Karayev, J.~Long, R.~Girshick,
  S.~Guadarrama, and T.~Darrell, ``Caffe: Convolutional architecture for fast
  feature embedding,'' in \emph{Proceedings of the ACM International Conference
  on Multimedia}, 2014.

\bibitem{everingham2010pascal}
M.~Everingham, L.~Van~Gool, C.~K. Williams, J.~Winn, and A.~Zisserman, ``The
  pascal visual object classes (voc) challenge,'' \emph{International Journal
  of Computer Vision}, vol.~88, no.~2, pp. 303--338, 2010.

\bibitem{riche2013saliency}
N.~Riche, M.~Duvinage, M.~Mancas, B.~Gosselin, and T.~Dutoit, ``Saliency and
  human fixations: State-of-the-art and study of comparison metrics,'' in
  \emph{Proceedings of the IEEE International Conference on Computer Vision},
  2013, pp. 1153--1160.

\bibitem{bruce2009saliency}
N.~D. Bruce and J.~K. Tsotsos, ``Saliency, attention, and visual search: An
  information theoretic approach,'' \emph{Journal of Vision}, vol.~9, no.~3,
  pp. 5--5, 2009.

\bibitem{zhang2013saliency}
J.~Zhang and S.~Sclaroff, ``Saliency detection: A boolean map approach,'' in
  \emph{Proceedings of the IEEE International Conference on Computer Vision},
  2013, pp. 153--160.

\bibitem{xu2014predicting}
J.~Xu, M.~Jiang, S. Wang, M. S. Kankanhalli, and Q. Zhao,
``Predicting human gaze beyond pixels,'' in
  \emph{Journal of vision}, vol.~14, no.~1, pp. 28-28, 2014.

\end{thebibliography}
\begin{IEEEbiographynophoto}
{Wenguan Wang} received the B.S. degree in computer science and technology from the Beijing Institute of Technology in 2013.
He is currently working toward the Ph.D. degree in the School of Computer Science, Beijing Institute of Technology, Beijing, China.
His current research interests include computer vision and deep learning.
He received the Baidu Scholarship in 2016.
\end{IEEEbiographynophoto}
\begin{IEEEbiographynophoto}
{Jianbing Shen} (M'11-SM'12) is a Full Professor with the School of
Computer Science, Beijing Institute of Technology, Beijing, China.
He has published about 100 journal and conference papers such as
\textit{IEEE TPAMI}, \textit{IEEE TIP}, \textit{IEEE TNNLS}, \textit{IEEE TVCG}, \textit{IEEE CVPR}, and \textit{IEEE ICCV}.
He has also obtained many flagship honors including the Fok Ying Tung Education Foundation from Ministry of Education,
the Program for Beijing Excellent Youth Talents from Beijing Municipal Education Commission, and the
Program for New Century Excellent Talents from Ministry of Education.
His current research interests are in the areas of Computer Vision for Autonomous Driving, Deep
Learning for Video Surveillance, Computer Graphics and Intelligent Systems.
He sevres as an Associate Editor for \textit{Neurocomputing} journal.
\end{IEEEbiographynophoto}

\vfill


\end{document}